\documentclass[runningheads,orivec]{llncs}

\usepackage{amsthm,amssymb,amsfonts}
\usepackage{algorithmic}
\usepackage{balance}
\usepackage{booktabs}
\usepackage{cite}
\usepackage[T1]{fontenc}
\usepackage[acronym]{glossaries}
\usepackage{graphicx}
\usepackage{textcomp}
\usepackage[dvipsnames,table,xcdraw]{xcolor}
\usepackage[caption=false,farskip=0pt,position=bottom]{subfig}
\usepackage{multirow}
\usepackage[hyphens]{url}
\usepackage[colorlinks=true,allcolors=black]{hyperref}
\usepackage[capitalise]{cleveref} %
\usepackage{xspace}
\usepackage[normalem]{ulem}
\usepackage{soul}

\DeclareMathOperator*{\argmax}{arg\,max}

\newcommand*{\RL}[2][]{\textcolor{Rhodamine}{[\textbf{\ifthenelse{\equal{#1}{}}{RL}{RL(#1)}}: #2]}}

\newcommand\customorcidAuthor[1]{\hspace*{-1mm}
\href{https://orcid.org/#1}{\includegraphics[width=0.3cm]{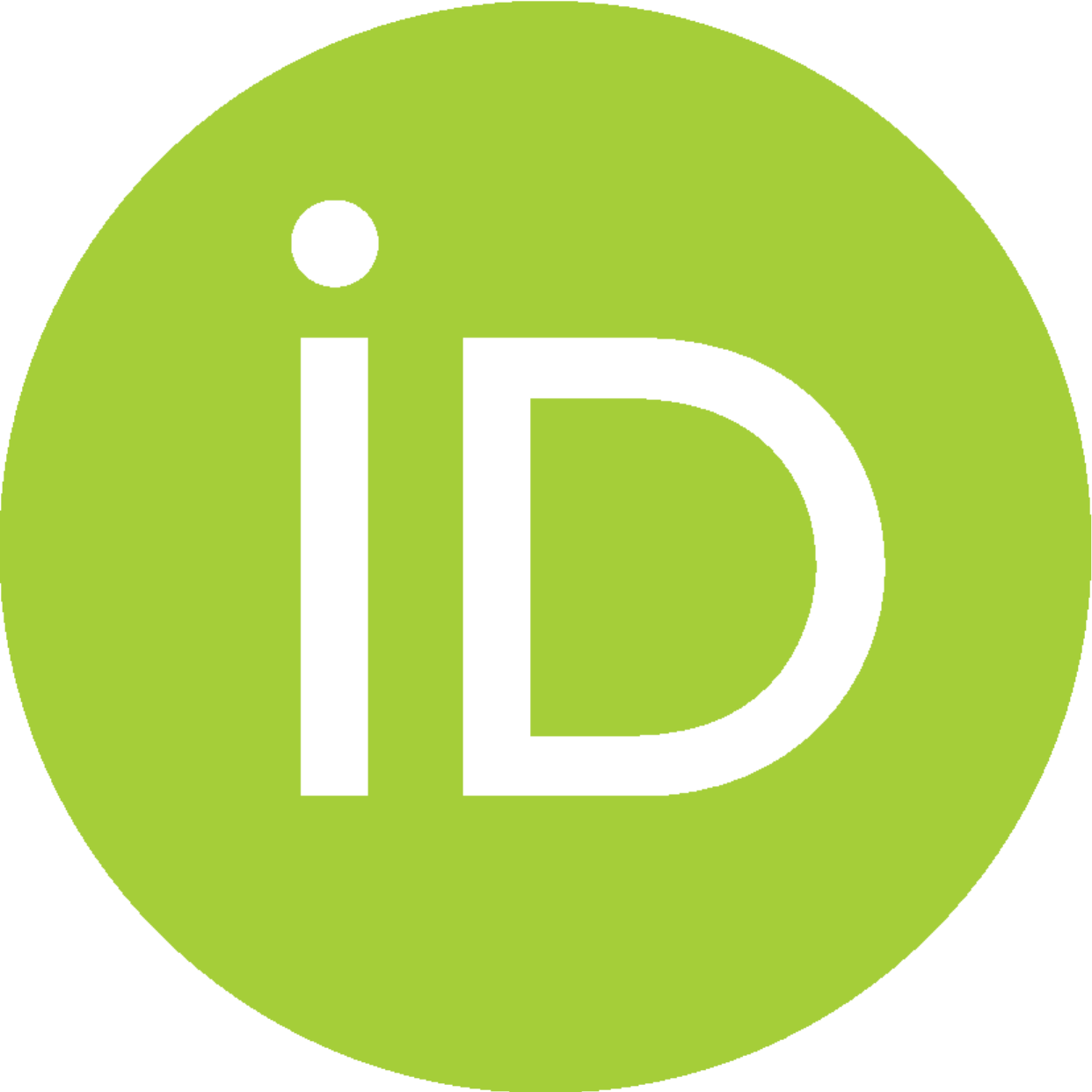}}\hspace*{-1mm}
}

\newacronymstyle{long-short-br}
{%
  \GlsUseAcrEntryDispStyle{long-short}%
}%
{%
  \GlsUseAcrStyleDefs{long-short}%
}
\setacronymstyle{long-short-br}

\newif\ifarxiv
\arxivtrue

\usepackage{transparent}
\usepackage{tikz}
\ifarxiv
    \newcommand\copyrighttext{%
      \scriptsize Accepted at ICPR 2026. The final published version is available via \textit{Springer} (DOI: \href{https://doi.org/10.1007/978-3-032-31927-2_42}{\textcolor{blue}{10.1007/978-3-032-31927-2\_42}}).}
    \newcommand\copyrightnotice{%
    \vspace{-5mm}
    \begin{tikzpicture}[remember picture,overlay]
    \node[anchor=south,yshift=80pt,xshift=0pt] at (current page.south) {\fbox{\transparent{0.85}\parbox{\dimexpr0.65\textwidth-\fboxsep-\fboxrule\relax}{\copyrighttext}}};
    \end{tikzpicture}%
    }
\else
\fi

\begin{document}

\tolerance=999
\sloppy

\newacronym{bisst}{Bi-SST}{bidirectional single-stream}
\newacronym{bert}{BERT}{Bidirectional Encoder Representations from Transformers}
\newacronym{bmt}{BMT}{Bi-Modal Transformer}
\newacronym{c3d}{C3D}{Convolutional 3D Network}
\newacronym{cezsar}{CEZSAR}{Contrastive Embedding Method for
Zero-Shot Action Recognition}
\newacronym{cnn}{CNN}{Convolutional Neural Network}
\newacronym{dap}{DAP}{Deep Action Proposal}
\newacronym{ed}{ED}{Elaborative Descriptions}
\newacronym{fps}{FPS}{frame per second}
\newacronym{gcn}{GCN}{Graph Convolutional Network}
\newacronym{glove}{GloVE}{Global Vectors}
\newacronym{gan}{GAN}{Generative Adversarial Network}
\newacronym{gpu}{GPU}{Graphics Processing Unit}
\newacronym{gru}{GRU}{Gated Recurrent Unit}
\newacronym{har}{HAR}{Human Action Recognition}
\newacronym{i3d}{I3D}{Inflated 3D Network}
\newacronym{idt}{IDT}{Improved Dense Trajectories}
\newacronym{lstm}{LSTM}{Long Short-Term Memory}
\newacronym{mdvc}{MDVC}{Multi-modal Dense Video Captioning}
\newacronym{mse}{MSE}{Mean Squared Error}
\newacronym{nlp}{NLP}{Natural Language Processing}
\newacronym{of}{OF}{Optical Flow}
\newacronym{rnn}{RNN}{Recurrent Neural Network}
\newacronym{sbert}{SBERT}{Sentence-BERT}
\newacronym{sota}{SOTA}{state of the art}
\newacronym{svm}{SVM}{Support Vector Machine}
\newacronym{tac}{TAC}{trimmed action classification}
\newacronym{vem}{VEM}{Visual Embedding Module}
\newacronym{sem}{SEM}{Sentence Embedding Module}
\newacronym{zsar}{ZSAR}{Zero-Shot Action Recognition}

\title{CEZSAR: A Contrastive Embedding Method for Zero-Shot Action Recognition}
\titlerunning{A Contrastive Embedding Method for Zero-Shot Action Recognition}

\author{Valter~Estevam\inst{1}$^,$\inst{2}\customorcidAuthor{0000-0001-9491-5882} \and
Rayson~Laroca\inst{2}$^,$\inst{3}\customorcidAuthor{0000-0003-1943-2711} \and\\
Helio~Pedrini\inst{4}\customorcidAuthor{0000-0003-0125-630X}\hspace{0.05mm} \and
David~Menotti\inst{2}\customorcidAuthor{0000-0003-2430-2030}}

\institute{Federal Institute of Paraná, Irati, Brasil \and
Federal University of Paraná, Curitiba, Brasil \and
Pontifical Catholic University of Paraná, Curitiba, Brasil \and
University of Campinas, Campinas, Brasil
}
\authorrunning{Estevam et al.}
\maketitle

\ifarxiv
    \copyrightnotice
\else
\fi

\begin{abstract}
This paper proposes a novel Zero-Shot Action Recognition~(ZSAR) method based on contrastive learning. In ZSAR, we aim to classify examples from classes that were missing during training. Two well-known problems remain in ZSAR: the semantic gap and the domain shift. A semantic gap occurs because label representations come from the textual domain (i.e., language models) and must be associated with visual representations (i.e., CNNs, RNNs, transformer-based). This multimodal nature implies that the semantic properties of the two spaces are not identical. On the other hand, the domain shift arises from differences between the training and test sets and is inherent to ZSAR once the test set is unknown. One of the most promising methods to address both issues is learning joint embedding spaces. Therefore, we propose a new model that encodes videos and sentences in a joint embedding space, trained by aligning videos with their natural-language descriptions. We design an automatic negative sampling procedure to augment the training dataset and generate unpaired data, i.e., visual appearance and unrelated descriptions. Our results are state-of-the-art on the UCF-101 and Kinetics-400 datasets under several split configurations. Our code is available at https://github.com/valterlej/cezsar.
\keywords{Semantic gap  \and Language-video representation \and Contrastive learning \and Zero-shot learning.}
\end{abstract}

\section{Introduction}
\label{sec:introduction}
\glsresetall

Zero-shot learning is a well-established problem in computer vision that aims to classify instances belonging to classes that were not available for training the models, usually called unknown or unseen classes~\cite{ma:2022prl}.
Nowadays, there are zero-shot approaches for objects~\cite{li:2022siamesecontrastive,chen:2024anydoor,sun:2024}, human actions~\cite{estevam:2022objsentzsar,gowda:2024continual,gowda:2023synthetic,huang:2022combtextimage}, and many other domains~\cite{radford:2021clip}.
This work focuses on~\gls*{zsar} in videos, i.e., in classifying instances (short video clips up to 10$\,$s duration) of unknown action classes. This particular problem has attracted the attention of the computer vision community in the last decade~\cite{estevam:2021survey}.

The most popular human action recognition approaches employ supervised learning, requiring a massive set of annotated videos for training, and keeping these models up to date is extremely challenging because new actions emerge every day as new objects, techniques, and forms of human interaction appear.
Moreover, new actions are rare and unavailable on YouTube or other large-scale sources. Even when available, the inclusion of new classes requires retraining existing models, demanding extensive computational resources, energy, and human labor to annotate instances with appropriate labels~\cite {estevam:2024tell}.

In~\gls*{zsar}, on the other hand, the need for annotations is transferred from the instances to the action classes.
It takes a lot less work to annotate classes (a few hundred annotations) than it does to annotate tens or hundreds of thousands of instances.
Hence, several pioneer works considered a set of attributes defined by humans as semantic information~\cite{liu:2011attr}. This representation is called prototype and ideally represents the archetype for each class. 
Nevertheless, even such an approach requires a lot of human effort and is not scalable, being replaced by an automatic procedure called label embedding, which uses word embedding methods~\cite{wang:2017bidilel} or sentence embedding methods~\cite{chen:2021,estevam:2024tell} to define the prototypes.
Recently, the most promising \gls*{zsar} methods relate visual appearance (e.g., given by some neural network) with semantic class information associated with their label projecting them into a joint embedding space. Due to the multi-modal nature, two crucial problems remains in these approaches: the domain shift and the semantic gap between the~modalities. 

\begin{figure*}[!htb]
	\centering
	\captionsetup[subfigure]{captionskip=-0.25pt,font={scriptsize},justification=centering}
	\vspace{0.75mm}
	\resizebox{0.995\linewidth}{!}{
            \subfloat[][ZSARCAP~\cite{estevam:2024tell}~(Acc. 67.2\%)\label{fig:tsnemotivation:a}]{
			\includegraphics[width=0.45\linewidth]{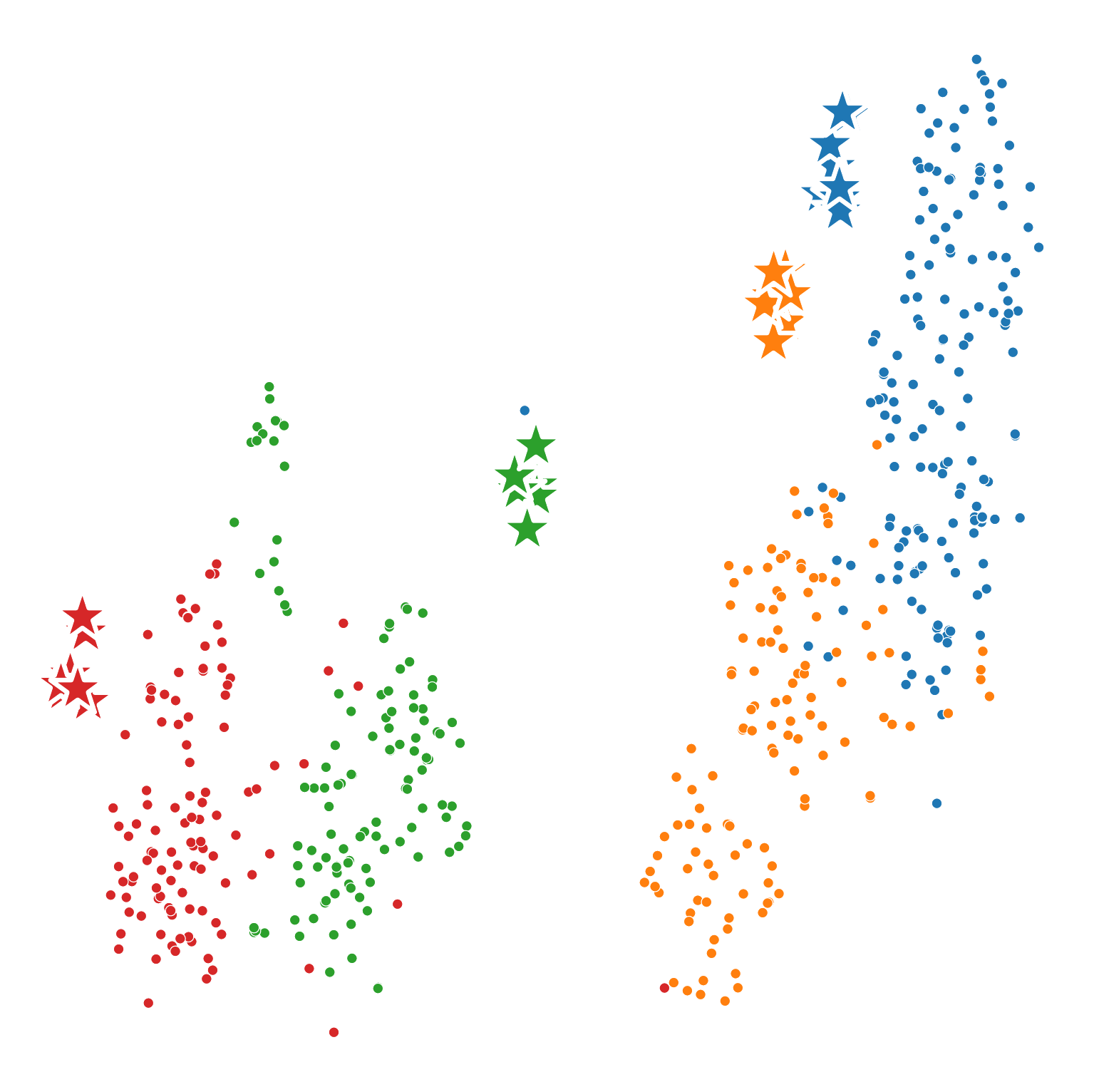} 
		} \, \textcolor{gray}{\vline} \,
            \subfloat[][Ours (Acc. 95.2\%)\label{fig:tsnemotivation:b}]{
			\includegraphics[width=0.45\linewidth]{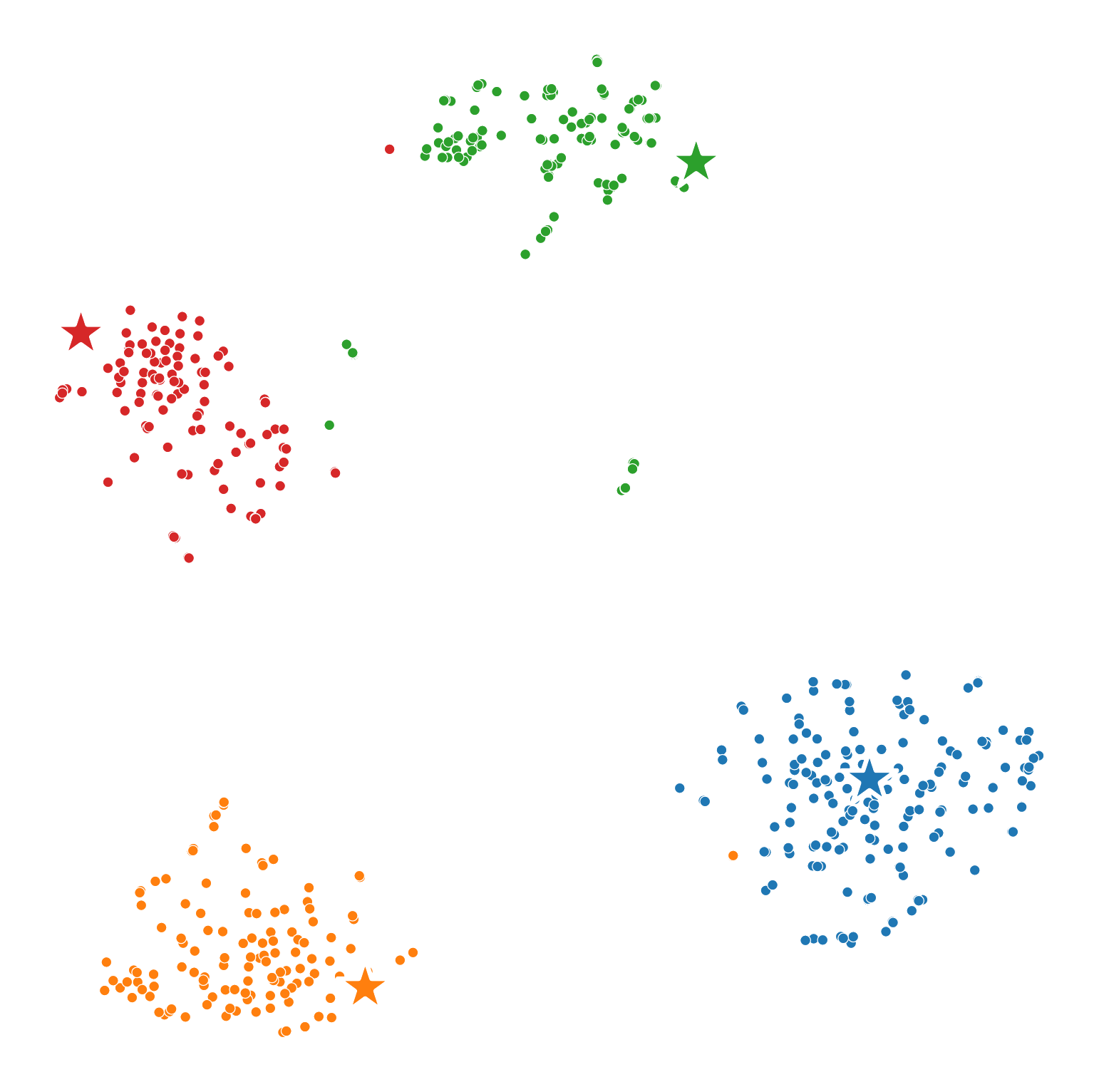} 
		}
	}    	    
	\vspace{-1mm}
    \caption{
    T-SNE visualization for a subset with the classes Horse Riding (blue), Horse Race (orange), Pommel Horse (green), and Balance Beam (red). 
    Dots are videos, and stars are label~prototypes.
    }
\label{fig:tsnemotivation}
\end{figure*}

The semantic gap is the information difference for each modality used by the methods, i.e., the distribution of instances in visual space is often distinct from that of their underlying semantics in semantic space~\cite{wang:2017bidilel}. For example, in \cref{fig:tsnemotivation:a}, we demonstrate that this problem occurs even in joint embedding-based models such as ZSARCAP~\cite{estevam:2024tell} or our proposed method.
The dots in the figure represent the video embeddings, and the stars the label embeddings. The lack of information and the challenges in relating them are the origins of this problem.
For instance, \textit{Pommel Horse} (green) and \textit{Balance Beam} (red) are usually performed in gymnasiums.
Therefore, they present similar frames in which the scene structure is similar, only differing in the artistic gymnastic equipment and some specific motions.

A strategy to mitigate the semantic gap on the visual side is to provide temporal information to learn a motion signature~\cite{wang:2017bidilel}. 
Another strategy is to explore the relationships among actions and objects~\cite{mettes:2021,estevam:2022objsentzsar}. 
These relationships occur in videos and texts.
Thus, it is possible to recognize objects in scenes and infer the action because the same information is used.
This last approach is robust in visual-semantic representation but fails in temporal modeling, which is essential to recognize actions independent of scenarios or objects (e.g., \textit{run, turn, punch,} and \textit{head massage}).

The semantic gap is also present in label encoding. Methods extensively used, such as Word2Vec or GloVe, fail to capture fine-grained differences because they project similar concepts (e.g., \textit{Horse Riding} and \textit{Horse Race}) close and, in some cases, also dissimilar ones (e.g., \textit{Pommel Horse and Horse Riding}).
Moreover, the label encoding process usually produces one array\footnote{This array represents the class prototype.}~for which we assume all required semantic information is encoded. Strategies to alleviate the semantic gap in label encoding include adding more descriptive texts and using better encoders, such as LLM-based encoders.~\cite{chen:2021,estevam:2024tell,estevam:2022objsentzsar}.
As shown in \cref{fig:tsnemotivation:b}, our method generates a better separation among the classes (for both videos and prototypes) and a lower distance between prototypes and their corresponding~videos.

Even though we have good descriptors for videos and texts, the domain shift problem remains unsolved. It corresponds to the differences in the probability distribution for the patterns in the training set compared to the test set~\cite{wang:2017bidilel}.
We believe that textual semantics is much less affected by domain shift than visual. Hence, learning a joint embedding space for these modalities, conditioned by textual descriptions, should alleviate the domain shift problem for visual patterns and reduce the semantic gap between information modalities.
Taking this into account, we propose a new method for \gls*{zsar}, called \gls*{cezsar}.
It consists of a joint projection method trained with an additional dataset containing untrimmed videos paired with human-generated sentences describing what is occurring in the~videos. 

\begin{figure}
\centerline{\includegraphics[width=0.875\columnwidth]{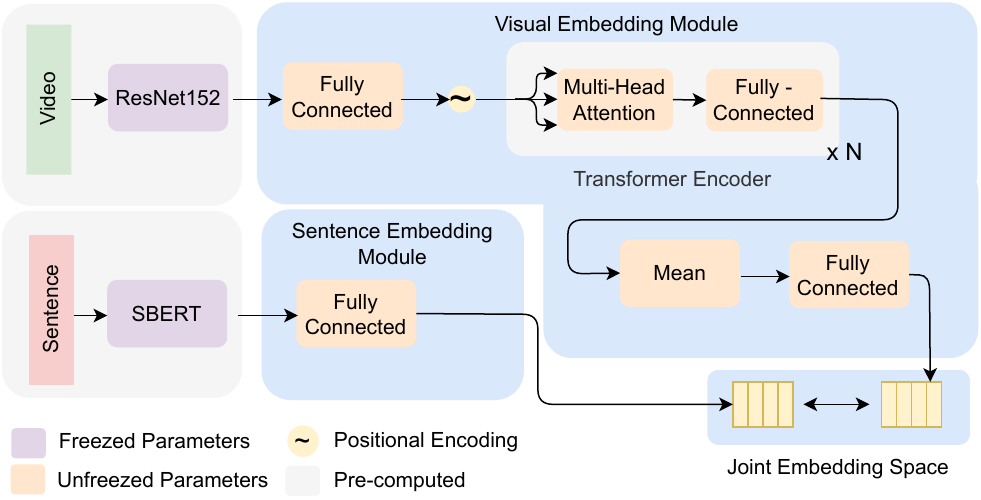}}
\vspace{-2mm}
\caption{
Our method is composed of the Visual Embedding and Sentence Embedding modules.
Each module produces a dense representation that is expected to be close if the sentence describes the video and far~otherwise.
}
\label{fig:proposedmethod}
\end{figure}

As illustrated in \cref{fig:proposedmethod}, 
our proposed model is a neural network with two modules.
The first, called~\gls*{vem}, is responsible for encoding visual information given by a pre-trained \gls*{cnn} (e.g., ResNet-152).
In this module, the videos are sampled at 1~\gls*{fps} and passed through the \gls*{cnn}, resulting in a feature stack.
There is also a fully connected layer responsible for reducing the stack dimensionality and feeding a Transformer Encoder.
This encoder uses self-attention to model temporal information for the videos.
Therefore, we have two dense representations for which we expect to be close if the text describes the video and distant otherwise.

For training the model, we propose a hard negative sampling method.
This method seeks negative alignments between videos and texts without human supervision. Thus, we can generate triplets (video, positive description, and negative description) and employ a triplet loss function. 
Our training process does not require a closed set of classes, but only enough pairs of videos and descriptions in natural language,
and it can be completed in just a few hours on a standard \gls*{gpu}. 

In summary, the main contributions of this work are: (i)~we introduce a new cross-modal contrastive learning method that effectively associates visual features and sentence descriptions with a reduced semantic gap; (ii)~our model enables projecting videos and descriptions with two distinct sub-networks. Hence, we can include additional information such as texts, images, or even videos; and (iii)~the robustness of our joint semantic space is demonstrated by reaching state-of-the-art results on the UCF-101 and Kinetics-400~datasets.

\section{Related Work}
\label{sec:related_work}

This section briefly discusses joint embedding learning employing sentences and contrastive learning.

\subsection{Joint Embedding Learning for ZSAR using Sentences}

Estevam~et al.~\cite{estevam:2024tell} proposed a method to represent both sides, videos and labels, with descriptive sentences.
They trained video captioning models~\cite{estevam2025dense} that produce a single sentence per video.
The video captioning method models temporal information in videos to infer probabilities over a vocabulary and generate a sentence. Although the results obtained with this technique were promising, there is much room for improvement in video captioning to effectively establish stronger associations between visual and textual patterns, which we believe would improve the performance of \gls*{zsar}. Subsequently, the same research group~\cite{estevam:2022objsentzsar} proposed to enrich the captioning sentences with textual descriptions given by objects recognized in the scenes, providing a robust set of semantic information that can be incorporated into our~model. 

\subsection{Contrastive Learning for Zero-Shot Learning}

Contrastive learning is a self-supervised learning technique that aims to learn a dense representation given label-visual pairs.
In learned space, similar pairs stay close together, and dissimilar pairs stay far apart.
Chopra~et al.~\cite{chopra:2005contrastive} were among the pioneers to propose a loss function for this problem.
Recently, Han~et al.~\cite{han:2021cegzsl} employed contrastive learning for generalized zero-shot learning, i.e., a sub-variant of the zero-shot learning problem that assumes the presence of seen and unseen classes in the test set. Although promising, their method was proposed for and evaluated on datasets that use manually annotated attributes to represent classes.

A benefit of contrastive learning is its robustness in preventing deep networks from overfitting noisy labels~\cite{xue:2022clbenefits}.
This property is critical for us because we deal with natural language descriptions that are intrinsically noisy due to ambiguities and annotators' perceptions of what should be described.
In addition, language-image pre-trained models such as CLIP~\cite{radford:2021clip} have attracted increasing attention from the research community~\cite{xu:2021videoclip,wu:2023bicrossmodal,lee:2024esczsar}.
These models have shown impressive results in zero-shot experiments, but they rely on extensive training infrastructure (e.g., clusters with up to $596$ Tesla V100 GPUs used for $18$~\cite{radford:2021clip}).
Moreover, the dataset containing $400$ million image-text pairs is not available for download, leading us to the following question: what would the results be for ER~\cite{chen:2021} or ZSARCAP~\cite{estevam:2024tell} trained with comparable infrastructure and a similar amount of data? Our proposed method, for example, has $1000\times$ fewer visual representations and $100\times$ fewer data pairs, is trained with $5\times$ less time on a single GPU, and achieves superior performance compared to non-clip-based methods.

\section{Classification Model}
\label{sec:classificationmodel}

\subsection{Problem Definition}

\gls*{zsar} can be stated as classifying a set of unseen action categories $ Z_{u} = \{z_{1},...,z_{u_{n}}\} $ (i.e., never seen before by the model).
It can be achieved by using a set of seen categories $ Z_{s} = \{z_{1},...,z_{s_{n}}\} $ so that $Z_{u} \cap Z_{s} = \emptyset$, or by transferring knowledge from other models trained without class labels, as in the proposed method.
As mentioned earlier, our model consists of a neural network compounded by two modules fed with pre-computed features for both modalities, visual and semantic description.
These modules are described in \cref{subsec:jointembeddermodel}.
As explained in \cref{subsec:contrastivelearning}, the model is trained in a contrastive way leveraging the proposed Hard Negative Sampling method, which is covered in \cref{subsec:hardnegativesampling}.
Finally, we present the~\gls*{zsar} procedure in \cref{subsec:zsarprocedure}.

\subsection{Joint Embedder Model}
\label{subsec:jointembeddermodel}

Initially, we explain the~\acrfull{vem}.
Given a video clip $v$ with $t$ seconds duration, we encode the frames at a rate of 1 FPS using a pre-trained \gls*{cnn}.
Then, we got $v_c \in \mathbb{R}^{t \times d_c}$, where $v_c$ is the feature stack for the video and $d_c$ is the \gls*{cnn} output dimension (e.g., using the ResNet-152 model $d_c=4.096$). 
This stack is fed to a fully connected layer aiming to reduce the dimensionality
\begin{equation}
\label{eq:dimensionalityreduction}
v_r = \text{ReLU}(v_{c}W + b) \, ,
\end{equation}
\noindent where $\text{ReLU}$ is a usual Rectified Linear Unit, $W$ is an internal weight matrix, $b$ is a bias~vector, and $v_r$ is the video stack projection into a lower dimensional space.
This stack is fed to a Transformer encoder, and the position of each feature is encoded with sine and cosine at different frequencies, as proposed by Vaswani~et al.~\cite{vaswani:2017attention}.
Then, these representations are passed through a multi-head attention layer that employs the scaled dot-product, defined in terms of queries ($Q$), keys ($K$), and values ($V$) as
\begin{equation}
    \label{eq:attention}
    \textit{Att}(Q,K,V)=\textit{softmax}(\frac{Q.K^{T}}{\sqrt{d_{k}}})V \,.
\end{equation}

The multi-head attention layer is a concatenation of several heads ($1$ to $h$) of self-attention ($Q~=~K~=~V~=~v_{r}^{\textit{PE}}$) applied to the input projections as
\begin{equation}
    \label{eq:multihead}
    \text{MHAtt}(v_{r}^{\textit{PE}},v_{r}^{\textit{PE}},v_{r}^{\textit{PE}})=[head_{1},..., head_{h}]W^{0} \, ,
\end{equation}
\noindent where $\textit{head}_{i}=\textit{Att}(v_{r}^{\textit{PE}}W_{i}^{v_{r}^{\textit{PE}}},v_{r}^{\textit{PE}}W_{i}^{v_{r}^{\textit{PE}}},v_{r}^{\textit{PE}}W_{i}^{v_{r}^{\textit{PE}}})$, $v_{r}^{\textit{PE}}$ is the $v_{r}$ positional encoded, and $[\text{ }]$ is a concatenation operator.

Afterward, a fully connected feed-forward network $\text{FFN}(\cdot)$ is applied to each position separately and identically
\begin{equation}
    \label{eq:ffn}
    \text{FFN}(u) = \max(0, uW_{1}+b_{1})W_{2}+b_{2} \, ,
\end{equation}
\noindent resulting in $v_{r}^{\text{FFN}}$. These features are averaged and fed to a fully connected layer responsible for projecting the result onto the joint semantic space, with $d_{emb}$ dimensions\footnote{In our experiments, $d_{emb}=128$.}, as
\begin{equation}
\label{eq:visjointprojection}
v_{\textit{emb}} = \text{ReLU}(\overline{v_{r}^{\text{FFN}}}W + b) \, .
\end{equation}

The~\acrfull{sem} takes a sentence $s$ and computes their \gls*{sbert}~\cite{reimers:2019sentencebert} representation $\text{SBERT}(\cdot)$, resulting in an array of $768$ dimensions.
This representation is fed to a fully connected layer to project onto the joint semantic space with $d_{emb}$ dimensions
\begin{equation}
\label{eq:sentjointprojection}
s_{\textit{emb}} = \text{ReLU}(\text{SBERT}(s)W + b).
\end{equation}

\subsection{Contrastive Learning and Loss Function}
\label{subsec:contrastivelearning}

We train our model using contrastive learning.
Our goal is to learn representations for which the video and its positive description are close to each other, and the video and its negative description are far apart.
Therefore, we employ the triplet loss~\cite{balntas:2016tripletloss} defined as
\begin{equation}
    \label{eq:tripletloss}
    \max(\parallel v_{\textit{emb}} - s_{\textit{emb}_{p}} \parallel - \parallel v_{\textit{emb}} - s_{\textit{emb}_{n}} \parallel + \epsilon, 0) \,,
\end{equation}
\noindent where $v_{\textit{emb}}$ is the output of our~\gls*{vem}, $s_{\textit{emb}_{p}}$ and $s_{\textit{emb}_{n}}$ are positive and negative sentence embeddings produced by our~\gls*{sem}, $||\cdot||$ is a distance metric, and $\epsilon$ is a margin ensuring that $s_{\textit{emb}_{p}}$ is at least $\epsilon$ closer to $v_{\textit{emb}}$ than $s_{\textit{emb}_{n}}$.

The positive description is annotated by humans, using natural language sentences. A complete description on how these annotations were made is available in~\cite{krishna:2017}.
As the dataset does not provide human-annotated negative samples, we design an automatic hard negative sampling procedure, described in the next~section.

\subsection{Hard Negative Sampling}
\label{subsec:hardnegativesampling}
Negative sampling is a straightforward procedure when the samples are class annotated.
We need to select samples from any other class randomly.
Similar samples can come from different classes, but human judgment is the ground truth. 
In our case, we have pairs of videos and descriptions, and using human judgment to evaluate the similarity degree of descriptions is infeasible.
Therefore, we employ a neural network ---~pre-trained in the paraphrasing task (i.e., the \gls*{sbert} model)~--- to evaluate if two different sentences have the same semantics.
We consider similar sentences if $\textit{Sim}(\text{SBERT}(x_{1}),\text{SBERT}(x_{2})) > 1 - \tau$\footnote{In our experiments, we set $\tau = 0.8$.}, where $\textit{Sim}$ is the cosine similarity.
We can find $n$ negative descriptions for each pair using this~rule\footnote{We set $n=10$.}. 

To improve our search for negative samples and augment the dataset, we evaluate the similarity of detected objects.
First, we filter two descriptive sentences for the detected objects most similar (using the rule previously defined) to the human-annotated sentence.
We then select a negative candidate for each positive description that is sufficiently different from each of these three positive descriptions (i.e., one from human annotation and two from object descriptions). 

Finally, for each temporal segment in the untrimmed videos, we randomly select three segments of up to 10 seconds. This augments the dataset by generating different positive pairs. Using these strategies, we obtained about three million triplets (video, positive description, and negative~description).

\subsection{ZSAR Classification}
\label{subsec:zsarprocedure}

Our classification consists of mapping both videos, including all semantic information available (i.e., visual and object definitions) and class semantic information (i.e., prototypes given by sentence class descriptions\footnote{More details in ~\cref{subsec:datasets}.}) into a joint embedding space. Then, the classification is performed with the nearest neighbor rule under some similarity function, such as
\begin{equation}
    \label{eq:zslclassification}
    z_{u_{pred}} = \argmax_{z_{u_{prot}} \in \mathcal{Z}_{u_{prots}}} \text{Sim}(\text{SE}(z_{u_{prot}}),\text{VidE}(v)) \,,
\end{equation}
\noindent in which $\text{Sim}$ is the cosine similarity; $v$ is a video, $z_{u_{prot}}$ is a sentence for each class,
$\text{SE}(\cdot)$ is a sentence embedding function defined in \cref{eq:sentjointprojection}, and $\text{VidE}(\cdot)$ is the video embedding function defined as
\begin{equation}
    \label{eq:videoembedding}
    \text{VidE}(v) = \alpha\text{VE}(v) + \beta\text{SE}(\text{O}(v))\,,
\end{equation}
\noindent where $\text{VE}(\cdot)$ is the visual embedding that uses the visual embedding module to encode the raw frames, $\textit{O}(\cdot)$ is responsible for encoding objects recognized in scenes with their definitions from WordNet (as in~\cite{estevam:2022objsentzsar}). The object classification can be performed directly from the ResNet-152 pre-computed features. Finally, $\alpha$ and $\beta$ control the importance of each semantic~feature.

\section{Datasets, Protocol and Implementation Details}
\label{subsec:datasets}

Our \gls*{zsar} experiments were carried out on the well-known UCF-101~\cite{soomro:2012} and Kinetics-400~\cite{carreira:2017} datasets.
UCF-101 has $13{,}320$ videos from $101$ action classes, with an average duration of $7.2$ seconds sampled at $25$ \gls*{fps}.
Kinetics-400 is much larger, comprising $306{,}245$~videos from $400$ action classes with at least $400$ clips each.
All videos have a duration of $10$ seconds and were collected from YouTube.
It should be noted that we obtained only $242{,}658$ clips (i.e., $\approx$ 80\%) of the original dataset because many videos are~unavailable\footnote{Removed or unavailable in our region.}.

The joint embedding model is learned with the ActivityNet Captions dataset~\cite{krishna:2017}.
It is a large-scale collection of YouTube videos, with temporal segments annotated and described by humans, each segment receiving one sentence of description. There are $20{,}000$ untrimmed videos, divided into training, validation, and test sets, with $50$/$25$/$25$\% of the videos. We obtained $\approx 12{,}000$ videos from the training and validation subsets in this work for the same reasons as in Kinetics-400.

The model was trained using an AdamW optimizer with $\text{lr}=1e-4$, $\beta_{1}=0.9$, and $\beta_{2}=0.99$, with a weight decay of $1e-5$.
A batch size of $128$ was employed.
Each video was encoded by obtaining a ResNet-152 feature per second (generating stacks with up to 15 features of 4096-d)\footnote{The features are available at https://1drv.ms/f/c/099341b05c7977d7/IgDTNPr-qsdDQowglu0TNlvmAWx3reSAAOiXI2-PvW4Oeio?e=4F52DG}.
Starting points for frame sampling were randomly selected.
The visual branch of the model employs a Transformer encoder with $\text{d\_model}$ = 512, $\text{layers}=1$, and $\text{heads}=2$. The training was performed with 25 epochs and early stopping set to 10. The \gls*{zsar} classification is performed by projecting videos and labels onto a common space where the nearest neighbor classifier with $k=1$ is used.

We evaluate our model on the UCF-101 dataset using the traditional protocol that randomly splits the dataset into seen and unseen classes ($50$\%/$50$\% - $50$ runs; $80$\%/$20$\% - $50$ runs, and $0$/$100$\% - $1$ run). We take only the test split because our joint embedding model is pre-trained on ActivityNet Captions, as described before. Therefore, in our case, the splits are $0$\%/$50$\%; $0$\%/$20$\%, and $0$/$100$\%.
Considering the Kinetics-400 dataset, we evaluate the performance adopting the same number of random classes from~\cite{mettes:2017,bretti:2021,mettes:2021,estevam:2022objsentzsar}~(i.e., $25$ - $50$ runs, $100$ - $50$ runs, and $400$ - $1$~run). 

The labels were represented by descriptive sentences generated using Google Gemini 3 (Pro version). The model was asked to create descriptions defining the action, the objects involved, and the locations where it usually occurs\footnote{The prompt used was: ``Write a descriptive sentence for each human action in the following list. Include a definition of the action as well as information about the objects and places where it is usually performed. Return the results in JSON format. The list of human actions is: <<dataset labels>>''.}.
All experiments were conducted on a computer with an AMD Ryzen 9 7950X 4.5 GHz CPU, 64 GB of RAM, and two NVIDIA RTX 5070 Ti GPUs (16~GB each).

\section{Results and Discussion}
\label{sec:experiments:results}

Table~\ref{tab:ucf101results} shows the results for the UCF-101 dataset. We highlight three sections in the table: the first, with a list of recent works; the second, with the performances of~\cite{estevam:2022objsentzsar} using only objects ($\text{O}$) and using objects and captions ($\text{O}+\text{C}$). 
This model was chosen because it also constructs a joint embedding space, but employing \gls*{sbert} exclusively; finally, we include our results using visual features (i.e., considering $\text{VidE}(v) = \text{VE}(v)$ in \cref{eq:videoembedding}) and using our complete model (considering $\alpha=0.8$, $\beta=0.2$ in \cref{eq:videoembedding}). These values were defined based on the dominance of visual features.

\begin{table}[!htbp]
\centering
\setlength{\tabcolsep}{6pt}
\caption{Results on the UCF-101 dataset reporting accuracy (\%) under different numbers of test classes. No classes were used for training. The best results are highlighted. $\text{VE}$ = visual features; $\text{O}$ = objects; $\text{C}$ = captions.}
\vspace{-2mm}
\resizebox{0.75\linewidth}{!}{
\begin{tabular}{lcccc}
\toprule
\multirow{2}{*}{Model}  & \multirow{2}{*}{Classes}  & \multicolumn{3}{c}{UCF-101 -- Test classes} \\ \cmidrule(lr){3-5}
    &          &  $101$      &  $50$                       & $20$         \\
\midrule
Mettes and Snoek~\cite{mettes:2017}%
& $-$      &  32.8         & 40.4~$\pm$~1.0            & 51.2~$\pm$~5.0 \\
Mettes~\textit{et al.}~\cite{mettes:2021}%
& $-$      &  36.3         & 47.3                      & 61.1           \\
Kim~\textit{et al.}~\cite{kim:2021}%
& 51       &  $-$          & 48.9~$\pm$~5.8            & $-$            \\
Chen and Huang~\cite{chen:2021}%
& 51       &  $-$          & 51.8~$\pm~$2.9            & $-$            \\
Brattoli~\textit{et al.}~\cite{brattoli:2020}%
& 664      &  39.8         & 48                        & $-$            \\
Huang~\textit{et al.}~\cite{huang:2022combtextimage}%
& 51    & $-$      & 46.4~$\pm$~3.1           & $-$            \\
Kerrigan~\textit{et al.}~\cite{kerrigan:2021}%
& 664    &  40.1         & 49.2                      & $-$            \\
Doshi~\textit{et al.}~\cite{doshi:2024multimodal}
&  595   &  45.0         & $-$                       & $-$            \\
Huang~\textit{et al.}~\cite{huang:2023enhancing} 
&   51   &  $-$          & 45.9~$\pm$~3.4            & $-$            \\
Estevam~\textit{et al.}~\cite{estevam:2024tell}%
& $-$    &  $-$          & 49.0~$\pm$~3.5            & $-$            \\
Lin~\textit{et al.}~\cite{lin:2022crossmodal}%
& 664    &  $-$          & 58.7~$\pm$~3.3            & $-$            \\
Lin~\textit{et al.}~\cite{lin:2022crossmodal}%
& 605    &  46.7         & 55.9                      & $-$            \\
Gowda~\textit{et al.}~\cite{gowda:2022claster}%
&  51        & $-$       & 53.9~$\pm$~2.5            & $-$            \\
\midrule\midrule
Estevam~\textit{et al.}~\cite{estevam:2022objsentzsar} ($\text{O}$)%
& $-$ &  39.8 & 49.4~$\pm$~4.0   & 60.0~$\pm$~8.5  \\
Estevam~\textit{et al.}~\cite{estevam:2022objsentzsar} ($\text{O}+\text{C}$)%
& $-$ & 40.9 & 53.1~$\pm$~3.9 & 63.7~$\pm$~8.3 \\
\midrule\midrule
Ours (VE) & $-$ & 49.8 & 59.1~$\pm$~2.7 & 72.6~$\pm$~5.9 \\ %
Ours (VE + O) & $-$ & \textbf{50.9} & \textbf{59.4~$\pm$~3.6} & \textbf{72.9~$\pm$~5.9} \\ %
\bottomrule
\end{tabular}
}
\label{tab:ucf101results}
\end{table}

Under the $0$/$101$ configuration, we observe an expressive increment of $10.0$ percentage points~(p.p.) in accuracy compared to Estevam et al.~\cite{estevam:2022objsentzsar} and $4.2$ p.p. compared to Lin~et al.~\cite{lin:2022crossmodal}.
Even when using only visual features, our model performs better than the one presented in~\cite{lin:2022crossmodal}. It is worth noting that Lin et al.~\cite{lin:2022crossmodal} use 50\% of the UCF-101 classes, in addition to 605 classes from Kinetics-700, thereby significantly increasing the availability of training data. Our results on UCF-101 have marginally improved with the inclusion of object-level semantic~information.

Considering the experiments in the Kinetics-400 dataset shown in Table~\ref{tab:kineticsresults}, we reached better results than the~\gls*{sota} under all configurations. Semantic information was responsible for consistent improvements, strongly suggesting that the semantic gap has been reduced.
Comparing VE against VE + O in the $0$/$25$ configuration, we do not observe a real gain in mean accuracy from including objects, and the standard deviation has increased compared to results using only visual features.
On the other hand, under the $0$/$400$ configuration, the increase of 2.3 p.p. is significant due to the higher amount of unknown classes and high intra-class similarity in this dataset (e.g., eating: \textit{burger, cake, carrots, chips, doughnuts, hotdog, ice cream, spaghetti,} and \textit{eating~watermelon}).

\begin{table}[!htbp]
\centering
\setlength{\tabcolsep}{6pt}
\caption{Results on Kinetics-400 reporting accuracy (\%) under different numbers of test classes. No classes were used for training. The best results are highlighted. $\text{VE}$ = visual features; $\text{O}$ = objects; $\text{C}$ = captions.}
\vspace{-2mm}
\resizebox{0.675\linewidth}{!}{
\begin{tabular}{lccc}
\toprule
\multirow{2}{*}{Model}                                                             & \multicolumn{3}{c}{Kinetics-400 -- Test classes}                     \\ \cmidrule(lr){2-4}
                                                                                   & $400$           & $100$                   & $25$                    \\
\midrule
Mettes and Snoek~\cite{mettes:2017}%
& 6.0             & 10.8~$\pm$~1.0          & 21.8~$\pm$~3.5          \\
Mettes~\textit{et al.}~\cite{mettes:2021}%
& 6.4             & 11.1~$\pm$~0.8          & 21.9~$\pm$~3.8          \\
Bretti and Mettes~\cite{bretti:2021}%
& 9.8             & 18.0~$\pm$~1.1          & 29.7~$\pm$~5.0          \\
\midrule\midrule
Estevam~\textit{et al.}~\cite{estevam:2022objsentzsar} (O)%
& 20.4            & 32.4~$\pm$~2.4          & 49.3~$\pm$~6.8          \\
Estevam~\textit{et al.}~\cite{estevam:2022objsentzsar} (O + C)%
& 19.4            & 35.1~$\pm$~2.4          & 54.6~$\pm$~6.1          \\
\midrule\midrule
\textbf{Ours (VE)}                                & 21.5   & 38.4~$\pm$~2.0 & 58.1~$\pm$~4.6 \\ %
\textbf{Ours (VE + O)}                                & \textbf{23.8}   & \textbf{40.4~$\pm$~1.7} & \textbf{58.7~$\pm$~5.9} \\ %
\bottomrule
\end{tabular}
}
\label{tab:kineticsresults}
\end{table}

We investigated the impact of adjusting the parameter $\tau$ in the hard negative sampling procedure.
This parameter regulates the criterion for determining whether a sentence is a positive or negative example.
Essentially, $\tau$ is used in the unsupervised identification of pairs of negative sentences.
Higher values of $\tau$ make it easier for the model to find negative examples.
We found that it is challenging to mine negative examples when $\tau \leq 0.7$ and that the model fails to find a sufficient number of negative examples when $\tau \leq 0.6$.
Consequently, we restricted our analysis to the threshold values of $0.7$, $0.8$, and $0.9$ for distinguishing positive and negative examples.
The models were evaluated using the UCF-101 dataset with all its classes and employing \cref{eq:videoembedding} to encode the videos.
We observed that using a more relaxed criterion ($\tau~=~0.7$) to determine negative examples did not lead to better \gls*{zsar} classification results (47.8\% accuracy).
Moreover, restricting the definition of similar examples ($\tau~=~0.9$) did not lead to higher ZSAR performance (48.3\% accuracy).
Therefore, we ultimately adopted the threshold of $0.8$ in our experiments.

\begin{figure}[!ht]
	\centering 
	\resizebox{0.735\linewidth}{!}{
        \subfloat[][Acc. $47.4$\%]{
			\includegraphics[width=0.95\linewidth]{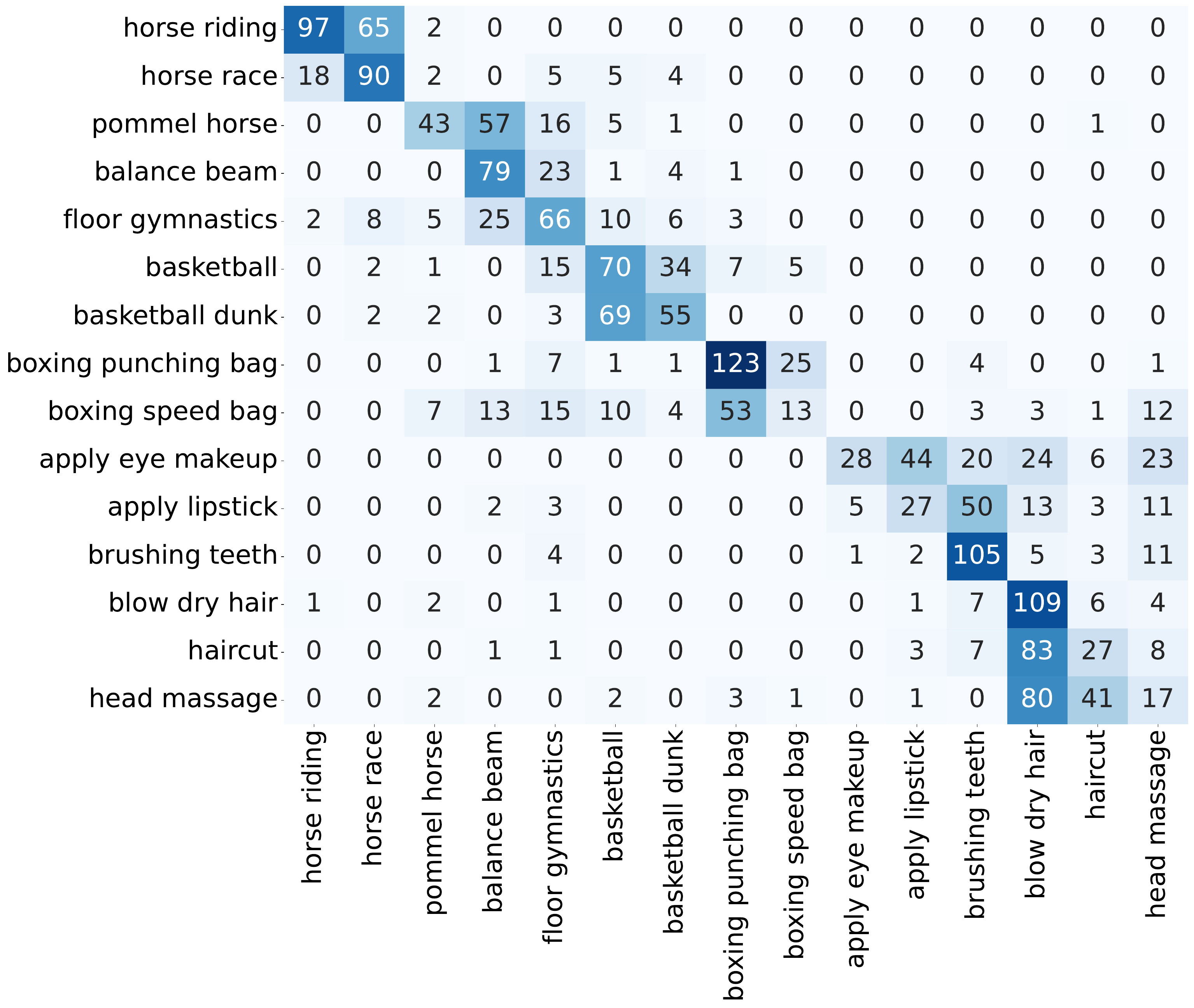}
		}  	
	}
    
    \vspace{5.0mm}
    
	\resizebox{0.735\linewidth}{!}{
            \subfloat[][Acc. $63.3$\%]{
			\includegraphics[width=0.95\linewidth]{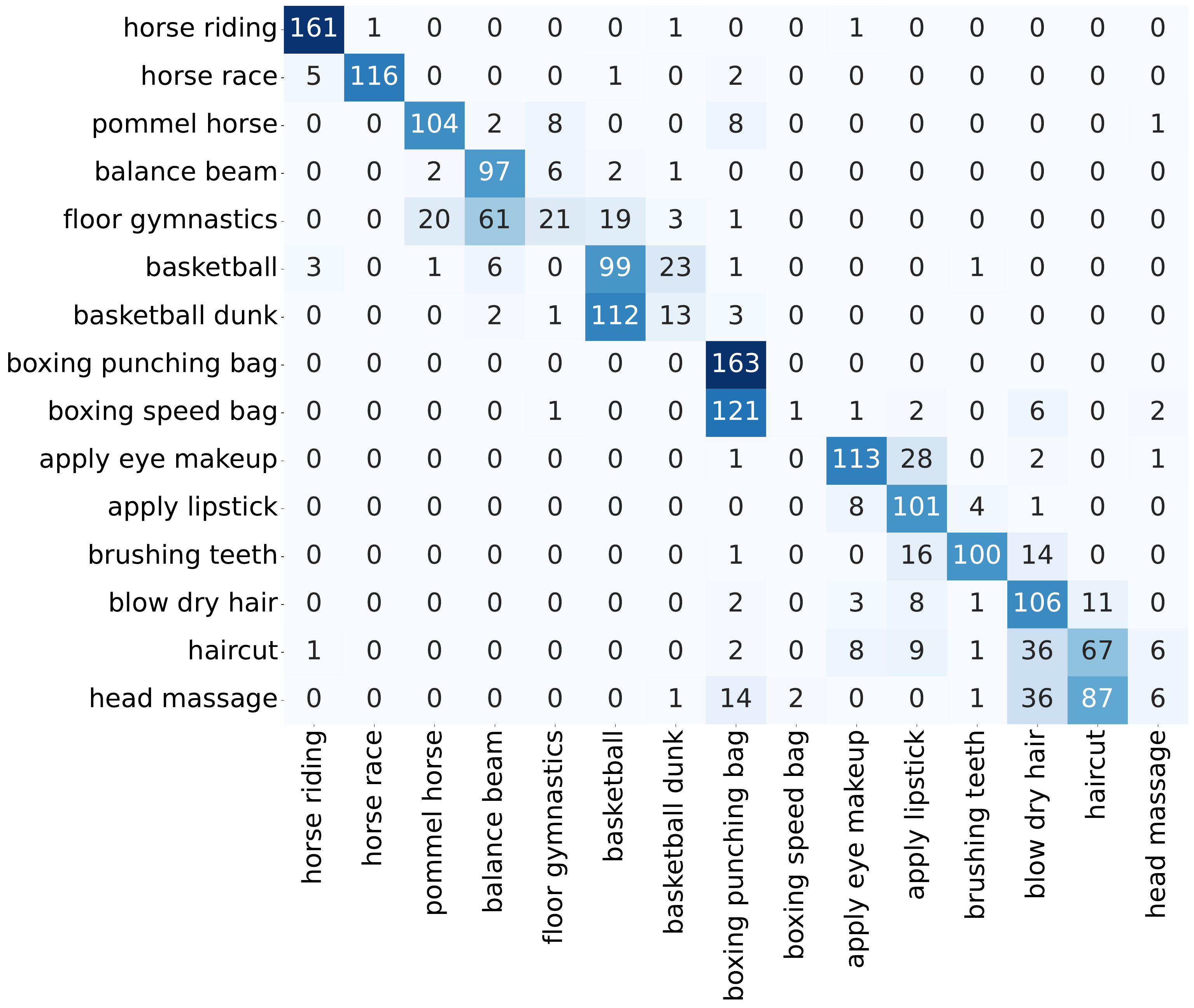} 
		}
	}

    \vspace{-1mm}
    
    \caption{(a) ZSARCAP~\cite{estevam:2024tell} results encoded with SBERT; (b) CEZSAR ($\textit{VE}+\textit{O}$).}
\label{fig:confusionmatrix}
\end{figure}

Finally, we evaluated our model on its ability to classify hard samples.  We compare our model against ZSARCAP~\cite{estevam:2024tell} because both use the same set of features as input and the same training dataset (i.e., ActivityNet Captions).
We observe an elevated increment in accuracy arising from improving the visual descriptor.
This is an excellent indication that, as shown in \cref{fig:tsnemotivation}, our joint space can approximate visual features of their semantic descriptions, narrowing the semantic gap and making the visual features less subject to domain~shift.
To investigate in more detail the relationship between the information modalities and the semantic gap, we choose a subset of $15$ classes from UCF-101 that are hard examples due to their high intra-class similarity.
These classes can be divided into six groups: (1 -- using horses) \textit{horse riding}, and \textit{horse race}; (2 -- performing gymnastics) \textit{pommel horse}, \textit{balance beam}, and \textit{floor gymnastics}; (3 -- using basketballs) \textit{basketball} and \textit{basketball dunk}; (4 -- boxing) \textit{boxing punching bag} and \textit{boxing speed bag}; (5 -- involving the face) \textit{apply eye makeup}, \textit{apply lipstick}, and \textit{brushing teeth}; (6 -- involving the hair) \textit{blow dry hair}, \textit{haircut}, and \textit{head massage}.
This subset is particularly hard because $30$ random runs of $15$ classes get $70.4\pm6.3$\% of accuracy against $63.3$\% (our model with the 15 selected classes).
\cref{fig:confusionmatrix} shows the confusion matrices for this subset.

When comparing each group's results for ZSARCAP and our method, we observed a reduction in confusion for all groups except 3 and 4. In group 3, our model was prone to classify all samples as basketball. On the other hand, in group 4, our model tends to classify \textit{boxing} videos as \textit{boxing punching bag}. We believe the inclusion of object semantics was not beneficial in these cases, although, overall, the results have been considerably better ($63.3$\% against $47.4$\%). Demonstrating that the most significant performance gain came from an effective reduction in the semantic gap and not just from the inclusion of more semantic~information.

\clearpage
\section{Conclusions}

Our conclusions are threefold:
(i)~contrastive learning is a straightforward yet effective approach for bridging the semantic gap between different information modalities in \gls*{zsar}. Comparing our approach against other architectures or training schemes, our results demonstrated a significant reduction in the semantic gap while allowing for easy inclusion of semantic information from other approaches without the need to retrain the contrastive model;
(ii)~conditioning the learning of visual features to a modality that is less impacted by the problem, such as texts, naturally reduces the domain shift problem. In all evaluated scenarios, the inclusion of text, even from a different domain, was beneficial to ZSAR performance and brought the samples closer to their corresponding prototypes;
and (iii)~automatic negative sampling is a practical method for augmenting a dataset without severely increasing the time required for the pre-computation of features, thus enabling training to be completed in just a few~hours on a single commercial GPU.
In future works, we intend to investigate the influence of the pre-training dataset size on contrastive learning performance and the impact of including images as label prototypes on the semantic gap.

\section*{Acknowledgments}

This study was supported in part by the \textit{Programa de Excelência Acadêmica~(PROEX) da Coordenação de Aperfeiçoamento de Pessoal de Nível Superior}~(CAPES), and by the \textit{Conselho Nacional de Desenvolvimento Científico e Tecnológico}~(CNPq) under grants \#~$315409$/$2023$-$1$ and \#~$304836$/$2022$-$2$. We also gratefully acknowledge the \textit{Pontifícia Universidade Católica do Paraná} and \textit{Fundação Araucária} for their financial support, which enabled conference participation. We further thank PROEQ-IFPR for providing the equipment used in this~research.

\bibliographystyle{splncs04}
\bibliography{references}

\end{document}